# Multi-Factors Aware Dual-Attentional Knowledge Tracing

Moyu Zhang, Xinning Zhu(✉), Chunhong Zhang, Yang Ji, Feng Pan and Changchuan Yin
Key Laboratory of Universal Wireless Communications, Ministry of Education
Beijing University of Posts and Telecommunications, Beijing, China
{zhangmoyu, zhuxn, zhangch, jiyang, pan_feng, ccyin}@bupt.edu.cn

## ABSTRACT

With the increasing demands of personalized learning, knowledge tracing has become important which traces students' knowledge states based on their historical practices. Factor analysis methods mainly use two kinds of factors which are separately related to students and questions to model students' knowledge states. These methods use the total number of attempts of students to model students' learning progress and hardly highlight the impact of the most recent relevant practices. Besides, current factor analysis methods ignore rich information contained in questions. In this paper, we propose Multi-Factors Aware Dual-Attentional model (MF-DAKT) which enriches question representations and utilizes multiple factors to model students' learning progress based on a dual-attentional mechanism. More specifically, we propose a novel student-related factor which records the most recent attempts on relevant concepts of students to highlight the impact of recent exercises. To enrich questions representations, we use a pre-training method to incorporate two kinds of question information including questions' relation and difficulty level. We also add a regularization term about questions' difficulty level to restrict pre-trained question representations to fine-tuning during the process of predicting students' performance. Moreover, we apply a dual-attentional mechanism to differentiate contributions of factors and factor interactions to final prediction in different practice records. At last, we conduct experiments on several real-world datasets and results show that MF-DAKT can outperform existing knowledge tracing methods. We also conduct several studies to validate the effects of each component of MF-DAKT.

## 1 INTRODUCTION

Knowledge Tracing (KT) is a crucial part of personalized learning, which can trace students' knowledge states. Specifically speaking, KT aims to predict students' responses to certain questions based on students' historical exercise sequences. By estimating students' knowledge states, KT can help students learn such as through recommending proper questions and learning materials [14], [27].

Up to now, there have been many researches in KT and current KT methods mainly involve three categories. The first category centers around Bayesian models such as Bayesian Knowledge Tracing (BKT) [23], [41]. BKT models students' knowledge states on a single concept as binary variables and uses a Hidden Markov Model (HMM) to update posterior distribution of binary variables. The second category centers around deep sequential models such as long-term memory networks (LSTM) [33]-related models [21], [28], [42] and Transformer-related models [11], [22].

The third category is factor analysis methods which utilize two kinds of factors separately related with students and questions to model students' learning progress. Item Response Theory (IRT) [29] uses the ability of students and characteristics of questions to analyze students' performance. Additive Factor Model (AFM) [1], [2] introduces a student-related factor which records students' attempts on relevant concepts to capture their historical practice behaviors. Performance Factor Analysis (PFA) [25] extends AFM by recording successful and failed attempts to differentiate the impact of practices with different outcomes. Knowledge Tracing Machine (KTM) [35] uses Factorization Machine (FM) [31] to model students' learning progress. KTM is quite a generic model that can incorporate all factors utilized in current factor analysis methods. DAS3H [5] uses KTM to extend DASH [16] which set time windows to count attempts of students to model forgetting behaviors. With the development of deep learning, both DIRT [4] and Neural Cognitive Diagnosis (NeuralCD) [36] have combined factor analysis methods with deep learning. They aim at cognitive diagnosis task which regards students' knowledge states as static and discover students' level of mastery on each concept. However, existing factor analysis methods still have some limitations. In this paper, we aim to deal with the following three main challenges:

First of all, current factor analysis models fail to highlight the impact of the most recent relevant practices of students. They mainly utilize the total number of students' attempts on concepts to denote students' historical practice behaviors. However, the most recent relevant exercises will have an important impact on students which is consistently agreed in educational psychology [9]. Although DAS3H sets time window to count the attempts, it fails to highlight the impact of the most recent practices because it gives practices in a time window the same weight. Besides, it's hard to select a proper window size. Secondly, current factor analysis methods ignore rich question information. Because students tend to have similar performance on similar questions and difficulty-level of questions also affect students' performance, we think relationship and difficulty level of questions should be considered. Thirdly, factor analysis models fail to differentiate contributions of factors and factor interactions in different practice records. In fact, factors and factor interactions should not equally contribute to the final prediction in different records [12].

Considering these aforementioned limitations of current factor analysis models, in this paper, we propose the Multi-Factors Aware Dual-Attentional Knowledge Tracing method (MF-DAKT).

First of all, to highlight the effect of the most recent relevant practice, we design a novel students-related factor called *recent factor* to record the most recent attempts of students on relevant concepts of target questions. Considering there may be a long-time interval between the most recent practice and target practice, we also apply a forgetting function to reduce the impact of distant relevant practices. The recent factor will be combined with other factors such as the total number of attempts by an attentional-convolution component to model students' learning progress.

Secondly, to enrich question representations, we introduce two kinds of questions' information including questions' relationship and difficulty to help construct a question graph as Figure 1 shows. Considering questions labelled with the same concepts are similar, some research define similarity between two questions as discrete values i.e., binary values, based on whether two questions have overlapped concepts [17], [40]. However, different numbers of overlapped concepts should lead to different similarities. We define the similarities as ***continuous values*** based on the number of overlapped concepts involved in two questions. What's more, there are research pointed out even two questions are labelled with the same concepts, they have individual difficulty [17]. Therefore, we can define difficulty level as the correctly answered rate of questions. At last, we use a pre-training method to enrich question representations based on both the relationship and difficulty level of questions. The pre-trained representations will be fine-tuned based on a regularization term about questions' difficulty level during the process of predicting students' responses on practices.

Thirdly, in order to differentiate contributions of each factor and factor interactions to final prediction in different records, and considering both factors and factor interactions are important, we design a ***dual-attentional knowledge tracing prediction method*** (DAKT). DAKT uses two sub-spaces and combines with attention mechanism to capture information contained in factors and factor interactions from different perspectives, respectively.

At last, we conduct experiments on several real-world datasets and results reveal that MF-DAKT can outperform the state-of-the-art KT models. Additionally, we perform several studies to demonstrate the effects of key components of MF-DAKT.

The contributions of our paper can be summarized as follows:
- As far as we know, we are the first to propose recent factor to capture the impact of the most recent attempt of students which can highlight the impact of the latest practices.
- To enrich question representations, we utilize a pre-training method to introduce two kinds of question information including questions' relationship and difficulty level. Pre-trained question representations will be fine-tuned based on a regularization term about the questions' difficulty level.

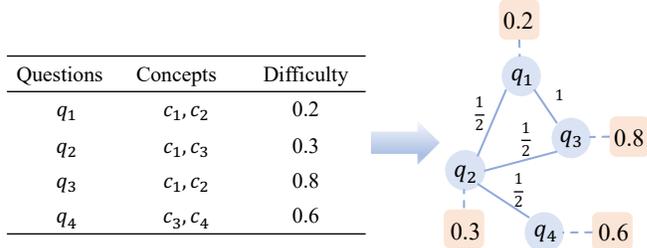

Figure 1: An example of question graph. $q_1$ and $q_3$ involve the same concepts and similarity is 1, and their difficulty levels are different. Both $q_1$ and $q_2$ involve two concepts and they only share one concept. Therefore, their similarity is $\frac{1}{2}$.

- Applying a dual-attentional mechanism, we not only can differentiate contributions of factors and factor interactions in different records, but also capture information contained in factors and factor interactions from different perspectives.

## 2 RELATED WORK

Recently, a lot of research have appeared in KT field. BKT [23], [41] is the most popular model which traces students' knowledge states of a certain skill by a hidden Markova model (HMM). It is the first model introduced into intelligent education field, and its variants introduce more actions such as slip and guess probability.

Later, with the development of deep sequential models, Deep Knowledge Tracing (DKT) [28] proposed to use LSTM to model students' knowledge states. Dynamic Key-Value Memory model (DKVMN) [42] extended DKT by utilizing a key-value memory network to discover relations between questions and underlying skills. DKT and DKVMN have been the benchmarks, but they use concepts to replace questions and fail to capture the individual differences among questions. To fully capture relations among concepts and questions, Graph-based Knowledge Tracing (GKT) [21], PEBG [17] and Graph-based Interaction model (GIKT) [40] proposed to construct a graph. GKT randomly initializes a concept graph and optimizes it by predicting students' performance which is computation-intensive and easily limited by the size of datasets. PEBG and GIKT define relations among questions based on concepts of questions. Besides, Relation-Aware Self-Attention model (RKT) [22] and Context-aware Attentive model (AKT) [10] use the attention mechanism to reduce the impact of irrelevant exercises on target questions and introduce decaying functions to model students' forget behaviors. RKT also integrates questions' textual content to capture question relations. AKT uses the IRT to capture differences **of** questions and avoid overparameterization.

In addition to Bayesian and deep sequential models, there is another type of methods in KT field, i.e., factor analysis methods, many of which are designed from the cognitive and psychometric perspectives such as IRT [29], AFM [1], [2] and PFA [25]. They consider factors which can affect students' knowledge states. These factors mainly center around students and questions such as the ability of students and the characteristics of questions. Some researches pointed out that traditional factor analysis methods can perform the same or even better than deep sequential KT models like DKT [37], [38], [39]. Multidimensional Item Response Theory (MIRT) [3] extend IRT to multidimension to capture more information contained in data. KTM [35] and DAS3H [5] incorporate most of the factors used in existing factor analysis methods and utilize FM [31] to model students' knowledge states. What's more, many variants of factor analysis models combine traditional factor analysis methods with deep learning methods. For example, DIRT [4] and NeuralCD [36] extend IRT to capture more complex information contained in data based on deep neural network, and ensure the interpretability of IRT at the same time.

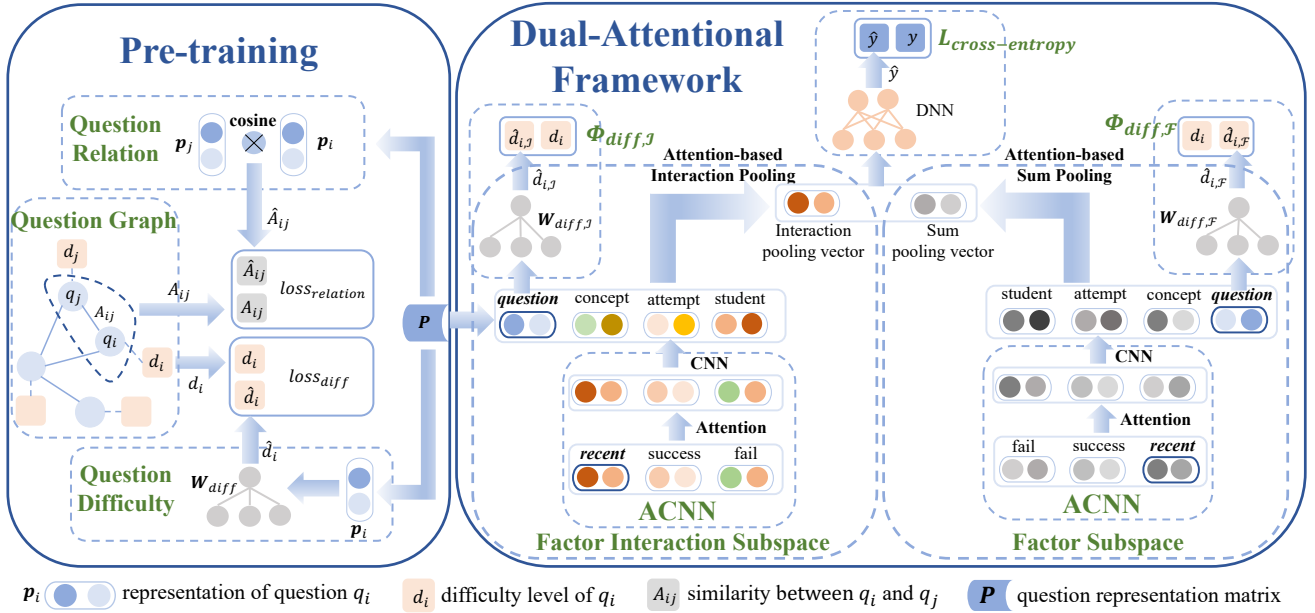

Figure 2: The architecture of Multi-Factors Aware Dual-Attentional Knowledge Tracing framework

## 3 PROPOSED METHOD

In this section, we introduce the details of our model MF-DAKT. The architecture of our models is shown in Figure 2. We propose a novel factor called recent factor to highlight the effect of the most recent relevant practice of students. Recent factor can be combined with other attempt-related factors which record the total number of students' attempts on relevant concepts to model students' learning progress based on an attentional-convolution component (ACNN). Moreover, we use a pre-training method to enrich question representations based on questions' relationship and difficulty level. The pre-trained question representations will be fine-tuned based on a regularization term about the questions' difficulty level during the process of predicting students' response on questions. At last, we apply a dual-attentional mechanism to differentiate the contributions of factors and factor interactions.

### 3.1 Problem Definition

Assume $U = \{u_i\}_{i=1}^{N_u}$ is the set of students, $Q = \{q_j\}_{j=1}^{N_q}$ is the set of questions and $C = \{c_k\}_{k=1}^{N_c}$ is the set of concepts involved in all questions. $N_u$, $N_q$ and $N_c$ are the number of students, questions and concepts, respectively. When students interact with online learning platforms, we can obtain students' historical interaction records $\chi_{u_i} = \{(u_i, q_1, y_1), \cdots, (u_i, q_{t-1}, y_{t-1})\}$, where $u_i$ denotes student $i$ and $y_{t-1} \in \{0, 1\}$ denotes whether $u_i$ answer question $q_t$ correctly at time $t-1$. Given a student's past exercise interactions before time $t$-1, we hope to predict the probability that student $u_i$ correctly answers questions $q_t$ at time step $t$ [7].

### 3.2 Input Factors

Factor analysis models consider two kinds of factors which center separately around students and questions. Current student-related factors mainly include the student, success and fail factors which separately denote the ability of students, the number of students' successful attempts and failed attempts, respectively. Question-related factors include question and concept factors which denote the characteristics of questions and relevant concepts of target question. In this paper, we also introduce a new student-related factor to record the most recent relevant practices of students. The encodings of commonly used traditional factors are as follow:

- **Student and Question Factors**. We can use one-hot vectors $u \in \mathbb{R}^{N_u}, q \in \mathbb{R}^{N_q}$ to represent students and questions.
- **Concept Factor**. Because one question is always related with multiple concepts in reality, we can use a multi-hot vector $c \in \mathbb{R}^{N_c}$ to represent concepts involved in a question.

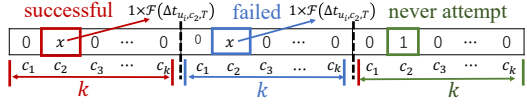

Figure 3: An example of success and fail factors of student $u_i$

Figure 4: An example of recent vector $r_{u_i, c_2, T}$

- **Success and Fail Factors**. We can separately count students' historical successful and failed attempts on the relevant concepts to represent students' learning process [35]. We separately record counts of successful and failed attempts of $u_i$ on relevant concepts of target question $q_T$ before time $T$ as two multi-hot vectors $s_{u_i,q_T,T}, f_{u_i,q_T,T} \in \mathbb{R}^{N_c}$. We also list an example of encoding success and fail factors in practice records of student $u_i$ as Figure 3 shows. The student $u_i$ firstly practice question $q_1$ which involves concept $c_1$ and $c_2$, and attempts on $c_1$ and $c_2$ are zero. When we want to predict his performance on $q_2$ which is also related with $c_1$ and $c_2$, failed attempts on $c_1$ and $c_2$ are all 1 and success attempts are all 0 because he answered $q_1$ wrongly before.

- **Recent Factor**. We can record the most recent practice of $u_i$ on concept $c_k$ at time $T$ as a one-hot vector $r_{u_i, c_k, T} \in \mathbb{R}^{3N_c}$. We set dimension as $3N_c$ to distinguish outcome of the last practice on concept $c_k$ of $u_i$ before time $T$. If the most recent practice on $c_k$ of $u_i$ is successful or failed, the $kth$ or $(N_c + k)th$ element of $r_{u_i, c_k, T}$ will be 1. If $u_i$ never practiced on $c_k$, $(2N_c + k)th$ element of $r_{u_i, c_k, T}$ will be 1. Considering students may practice relevant concepts a long time algo, we also introduce a forgetting formula $\mathcal{F}(\cdot)$. If students ever practiced on concept $c_k$, we record the time interval between the most recent practice on concept $c_k$ and current time step $T$, i.e., $\Delta t_{u_i,c_k,T}$. If $u_i$ ever practiced on concept $c_k$, the non-zero element of $r_{u_i, c_k, T}$ will be $1 \times \mathcal{F}(\Delta t_{u_i,c_k,T})$. By applying forgetting formula on $\Delta t_{u_i,c_k,T}$, we can reduce the influence of the most recent relevant practice. Figure 4 shows an example of $r_{u_i, c_2, T}$. Forgetting function can be various and we adopt a common approach [20], [30]:

$$\mathcal{F}(\Delta t) = e^{-\theta \cdot \Delta t} \quad (1)$$

where $\theta$ is a learnable decay rate. Considering decay rates on each concept may be different, we give each concept a unique decay rate in this paper:

$$\mathcal{F}(\Delta t_{u_i,c_k,T}) = e^{-\theta_{c_k} \cdot \Delta t_{u_i,c_k,T}} \quad (2)$$

Assume $C_{q_j}$ is the set of concepts related to target question $q_j$, we define recent factor as a multi-hot vector $r_{u_i,q_j,T}$:

$$r_{u_i,q_j,T} = \sum_{c_k \in C_{q_j}} r_{u_i,c_k,T}, r_{u_i,q_j,T} \in \mathbb{R}^{3N_c} \quad (3)$$

### 3.3 Pre-training Process

Current factor analysis methods learn question representations based on predicting students' performance and ignore question information. Here, we firstly introduce question information used in our model and then give a detailed description of pre-training.

- **Question Relation**. Each question is labeled with relevant concepts, and we can capture question relations as Figure 1 shows based on questions' concept labels. The relation in graph is motivated by intuition that questions sharing the same concepts can be similar. For example, "1+2" and "9+7" all involve "addition" concept and students who can answer "1+2" correctly are likely to answer "9+7" correctly, so we can think they are similar. A question may involve multiple concepts and different number of overlapped concepts should contribute to different similarities. Therefore, we define questions' similarities as continuous values as below:

$$A_{ij} = \frac{\left|C_{q_i} \cap C_{q_j}\right|}{\max\left(\left|C_{q_i}\right|, \left|C_{q_j}\right|\right)} \quad (4)$$

where $A_{ij}$ denotes the similarity between $q_i$ and $q_j$. $|C_{q_j}|$ is the number of concepts related to $q_j$. $|C_{q_i} \cap C_{q_j}|$ denotes the number of overlapped concepts between $q_i$ and $q_j$.

- **Question Difficulty.** Because questions related to the same concepts may have individual difficulty levels which have important impacts on learning outcomes [19], [24], we introduce difficulty levels of questions as a kind of question information. Usually, if a question's correctly answered rate is high, it can be considered to be easily to solved. Therefore, we can define the correctly answer rate of question $q_j$ in the training dataset as difficulty level of question $q_j$, i.e., $d_j$.

In this paper, we adopt a pre-training method to introduce questions' relationship and difficulty level to enrich question representations. More specifically, we can regard relations and difficulty levels of questions as two kinds of training targets that can be used to design the loss functions so as to enrich question representations as *Pre-training* component in Figure 2 shows.

To capture questions' relations, we use cosine similarity of two questions' representations to calculate their similarities as below:

$$\hat{A}_{ij} = \frac{p_i p_j}{||p_i||_2 ||p_j||_2} \quad (5)$$

where $p_j$ is the projected representation vector of question factor $q_j$. The projection is a parameter matrix $P \in \mathbb{R}^{N_q \times D}$, where $D$ is the hidden size of question representations. We adopt squared loss to enforce question representations to contain questions' relation:

$$loss_{relation} = \sum_{i=1}^{N_q} \sum_{j=1}^{N_q} (\hat{A}_{ij} - A_{ij})^2 \quad (6)$$

where $A_{ij}$ is the similarity label between $q_i$ and $q_j$.

Because questions also have individual difficulty levels, we utilize a fully connected layer to convert question representation vector into a scalar value, that is, difficulty level:

$$\hat{d}_i = p_i W_{diff} \quad (7)$$

where $W_{diff} \in \mathbb{R}^{D \times 1}$ is weight matrix of network layer. Similarly, we adopt a squared loss to measure error:

$$loss_{diff} = \sum_{j=1}^{N_q}(\hat{d}_i - d_i)^2 \quad (8)$$

where $d_i$ is the difficulty label of $q_i$.

To enrich question representations based on relationship and difficulty level of questions at the same time, we combine two loss functions mentioned above to jointly optimize parameters:

$L_{pre-train} = \lambda_1 loss_{relation} + \lambda_2 loss_{diff}, \{\lambda_1, \lambda_2\} \in [0,1] \quad (9)$

where $\lambda_1$ and $\lambda_2$ can flexibly control the impact on question representations of two kinds of question information.

### 3.4 Dual-Attentional Mechanism

Considering both factors and factor interactions are important, we apply a dual-attentional knowledge tracing method (DAKT) to differentiate contributions of factors and factor interactions in different practice records, and capture information contained in factors and factor interactions from different perspectives.

*3.4.1 Dual Subspaces.* DAKT adopts two subspaces including factor interaction and factor subspace to separately capture information contained in factor interactions and factors from different perspectives. Therefore, similar to multi-head attention mechanism [34], we project each factor to $D$ dimensions with two kinds of projections. We perform pre-training twice independently and obtain two question representations matrixes, i.e., $P_{\mathcal{F}}$ and $P_{\mathcal{J}}$, which project question factor into factor and factor interaction subspace, respectively. In addition to question factor, other factors such as student factor will be projected with randomly initialized parameter matrixes which will be optimized during the process of predicting students' performance. Because most of the operations in two subspaces are the same, we won't separately list the same operations of two spaces for the sake of simplicity. As for different operations in two subspaces, we will add two kinds of subscript on each operation to distinguish, where $\mathcal{F}$ denotes factor subspace and $\mathcal{J}$ denotes factor interactions subspace.

*3.4.2 ACNN Component.* Considering traditional factor analysis models fail to highlight the impact of the latest relevant practices of students, we propose recent factor which can be combined with success and fail factors to model students' learning progress based on an attentional-convolution component (ACNN). To reduce noise and highlight important factors, we firstly adopt an attention network. Mathematically:

$$B = (b_1, b_2, \cdots, b_{N_a}), B \in \mathbb{R}^{N_a \times D} \quad (10)$$
$$\alpha = Softmax(W_a^L(\cdots(W_a^1 B + b_1)) + b_L), \alpha \in \mathbb{R}^{N_a} \quad (11)$$

where $b_i$ is representation vector of $i$th factor. $N_a$ is the number of input attempt-related factors including success, fail and recent factors in this paper. $L$ is the number of attention layers in ACNN. The elements of $\alpha$ denotes the attention scores of input factors. We can give representations of factors a unique attention weight:

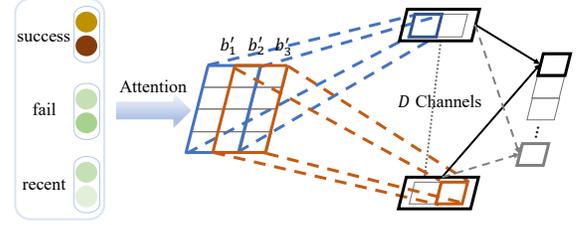

**Figure 5: Attentional Convolution Network Component**

$$B' = (b'_1, b'_2, \cdots, b'_{N_a}) = [b_i \alpha_i]_{i=1}^{N_a} \in \mathbb{R}^{N_a \times E} \quad (12)$$

We use reweighted representations in $B'$ to model students' learning progress. More specifically, we use a one-dimensional convolution neural network ($1d$ $CNN$) [15] which is an effective method to capture the interrelationships of the total attempts and attempts of recent practices as Figure 5 shows [26]. This operation can be simply defined as below:

$$A = 1d\ CNN(B') \quad (13)$$

Finally, we can obtain a vector $A \in \mathbb{R}^D$ called attempt vector which denotes historical practice behaviors of students and can be used to predict students' performance together with other factors.

*3.4.3 Attention-based Pooling.* Considering factors and factor interactions may have different contributions in different practice records. We can use an attention network to learn attention scores of factors. Attention and reweight operations are expressed as:

$$G = (g_1, g_2, \cdots, g_{N_p}), G \in \mathbb{R}^{N_p \times D} \quad (14)$$
$$\hat{\alpha} = Softmax(\widehat{W}_a^H(\cdots(\widehat{W}_a^1 G + \hat{b}_1)) + \hat{b}_H), \hat{\alpha} \in \mathbb{R}^{N_p} \quad (15)$$

where $N_p$ is the number of representation vectors of input factors in attention-based pooling component. Input vectors including projected representations of student, question, concept factors and the generated attempt vector $A$ from ACNN. $H$ is the number of attention layers. We can reweight representations of input factors:

$$G' = (g'_1, g'_2, \cdots, g'_{N_p}) = [g_i \hat{\alpha}_i]_{i=1}^{N_p} \in \mathbb{R}^{N_p \times D} \quad (16)$$

The obtained attention scores can also enhance interpretability to some extent because they can help us find out which factor is more important and help us understand outcome. At the same time, we find the attention scores can help DAKT specially model the factor interactions. The interaction can be expressed as $\langle g'_i, g'_j \rangle = \langle g_i, g_j \rangle \hat{\alpha}_i \cdot \hat{\alpha}_j$, where $\hat{\alpha}_i \cdot \hat{\alpha}_j$ can be regarded as the attention weight of interaction between $g_i$ and $g_j$.

**Two Kinds of Pooling Operations**. Similar to multiple subspaces in multi-head attention mechanism, above operations are performed in two subspaces in parallel. From two subspaces, we can get two kinds of representation vector sets, i.e., $G'_{\mathcal{F}} = (g'_{1,\mathcal{F}}, \cdots, g'_{N_p,\mathcal{F}})$ and $G'_{\mathcal{J}} = (g'_{1,\mathcal{J}}, \cdots, g'_{N_p,\mathcal{J}})$, which are separately obtained through operations of both *ACNN* component and *attention-based pooling* based on two kinds of projected dense representation vectors of factors in factor and factor interaction subspaces. $G'_{\mathcal{F}}$ and $G'_{\mathcal{J}}$ will be performed the interaction and sum pooling operations in two subspaces, respectively:

**Table 1: Dataset Statistics**

|  | ASSIST2009 | Algebra2006 | EdNet |
|---|---|---|---|
| students | 3,883 | 1,145 | 5,002 |
| questions | 17,737 | 129,255 | 10,779 |
| concepts | 123 | 493 | 188 |
| records | 282,668 | 1,817,016 | 222,141 |
| concepts / question | 1.20 | 1.01 | 2.21 |
| records / question | 15.94 | 14.06 | 20.61 |
| proportion of non-zero time interval | 87.67% | 95.15% | 69.15% |

$$\boldsymbol{V}_{\mathcal{F}} = \sum_{i=1}^{N_p} \boldsymbol{g}'_{i,\mathcal{F}}, \quad \boldsymbol{V}_{\mathcal{I}} = \sum_{i=1}^{N_p} \sum_{j=i+1}^{N_p} \boldsymbol{g}'_{i,\mathcal{I}} \odot \boldsymbol{g}'_{j,\mathcal{I}} \quad (17)$$

where $\odot$ denotes element-wise multiplication operation. $\boldsymbol{V}_{\mathcal{F}}$ and $\boldsymbol{V}_{\mathcal{I}}$ separately are sum pooling and interaction pooling vectors.

*3.4.3 Prediction Component.* At last, we concatenate two kinds of pooling vectors $\boldsymbol{V}_{\mathcal{F}}$ and $\boldsymbol{V}_{\mathcal{I}}$, and input them into a deep neural network to fully capture information contained in data and get the final prediction value. It can be abstracted as:

$$\hat{y} = DNN([\boldsymbol{V}_{\mathcal{F}}, \boldsymbol{V}_{\mathcal{I}}]) \quad (18)$$

To learn parameters in DAKT, we choose the common binary cross-entropy loss function to train model, i.e.,

$$L_{cross-entropy} = \sum_{i=1}^{|S|} y_i \log \hat{y}_i + (1-y_i) \log(1-\hat{y}_i) \quad (19)$$

where $S$ denotes the set of records for training. The pre-trained questions' representations will be fine-tuned in the KT prediction component. We also add the regularization terms about question difficulty level to restrict question representations to fine-tuning:

$$L_{total_{loss}} = L_{cross-entropy} + \Phi_{diff,\mathcal{F}} + \Phi_{diff,\mathcal{I}} \quad (20)$$

$$\Phi_{diff,\mathcal{F}} = \sum_{q_j \in S}^{S} (\boldsymbol{W}_{diff,\mathcal{F}} \boldsymbol{p}_{j,\mathcal{F}} - d_j)^2 = \sum_{q_j \in S}^{S} (\hat{d}_{i,\mathcal{F}} - d_j)^2 \quad (21)$$

$$\Phi_{diff,\mathcal{I}} = \sum_{q_j \in S}^{S} (\boldsymbol{W}_{diff,\mathcal{I}} \boldsymbol{p}_{j,\mathcal{I}} - d_j)^2 = \sum_{q_j \in S}^{S} (\hat{d}_{i,\mathcal{I}} - d_j)^2 \quad (22)$$

where $d_j$ is difficulty label of question $q_j$ and $\Phi_{diff,\mathcal{F}}, \Phi_{diff,\mathcal{I}}$ are regularization terms about difficulty level of two kinds of question representations in two subspaces. As section *3.4.1* mentioned, we perform pre-training twice and get two question representations matrixes $\boldsymbol{P}_{\mathcal{F}}$ and $\boldsymbol{P}_{\mathcal{I}}$. $\boldsymbol{p}_{j,\mathcal{F}}$ and $\boldsymbol{p}_{j,\mathcal{I}}$ separately are projected vectors of $q_j$ based on $\boldsymbol{P}_{\mathcal{F}}$ and $\boldsymbol{P}_{\mathcal{I}}$. $\boldsymbol{W}_{diff,\mathcal{F}}$ and $\boldsymbol{W}_{diff,\mathcal{I}}$ are the weight matrixes used to convert question representations into scalar values in two independent pre-training processes, respectively. $\hat{d}_{i,\mathcal{F}}, \hat{d}_{i,\mathcal{I}}$ separately are difficulty level calculated by $\boldsymbol{p}_{j,\mathcal{F}}$ and $\boldsymbol{p}_{j,\mathcal{I}}$.

## 4 EXPERIMENTS

In this section, we conduct a series of experiments on several real-world datasets to evaluate the performance of our model. We also show some ablation studies and visualizations to validate the impact of each key components in our MF-DAKT architecture.

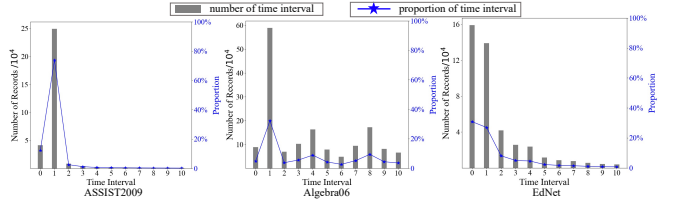

Figure 6: Statistics of time interval of the most recent practice

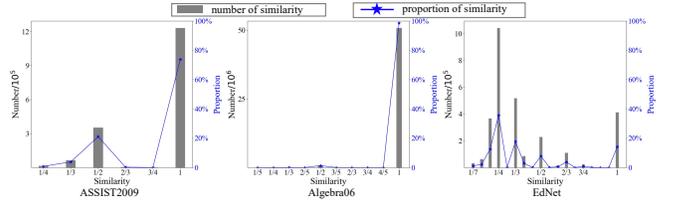

Figure 7: Statistics of similarities between questions

### 4.1 Datasets

We evaluate performance of MF-DAKT and several baseline KT models on predicting students' response using three KT datasets.

- **ASSISTments2009**. This dataset is one of the most popular benchmark datasets in KT field which is provided by the ASSISTments platform [9]. The version in our paper is conducted using the latest updated version "Skill-builder". There are 3,883 students, 17,737 questions and 123 concepts. We totally obtain 282,668 observed responses. What need to point out is that a question can involve 4 concepts at most.
- **Bridge Algebra2006**[1]. Bridge to Algebra 2006-2007 stems from the KDD Cup 2010 EDM Challenge [13]. There are 1,145 students, 129,255 questions and 493 concepts. We get in total 1,817,016 observed responses. For this dataset, we considered the combination of question and the step IDs as one answer which is recommended by challenge organizers. The largest number of concepts related with a question is 5.
- **EdNet**[2]. This dataset is collected by [6]. Because the whole dataset is too large, and we only choose EdNet-KT1 dataset which consists of students' practice history logs and randomly selected 222,141 records of about 5,002 students [17], [40]. There are in total 188 concepts and 10,779 questions. The largest number of concepts underlying one question is 6.

We excluded concept tags that are unnamed or dummies to remove the noise for three datasets and only use the sequence of students whose history records are longer than 3 because history exercise records that are too short are meaningless to predict. The complete characteristics of three datasets are summarized in Table 1. In this paper, we use two commonly used metrics for KT: AUC (Area Under the ROC curve) and ACC (Accuracy), where AUC is insensitive to class imbalance problem and is a more robust metric.

**Data Analysis.** To validate the necessity of introducing *recent factor* and defining similarities between questions as *continues*

---
[1] https://pslcdatashop.web.cmu.edu/KDDCup/downloads.jsp
[2] https://github.com/riiid/ednet

*values*, we conduct several data statistics on three datasets. We record the number of times that students have practiced relevant concepts before each target practice in students' historical practice sequence, and record corresponding time step intervals. For the convenience of observation, we only show the number of time interval within 10 time steps in Figure 6. If time interval is 0, it means when students practiced target questions, they had never attempted relevant concepts. As Figure 6 shows, the number of non-zero time intervals account for a large proportion of practice records. We also record the proportion of the number of non-zero time interval of total records in Table 1 which show that students often practice relevant concepts of target questions. This statistic can validate the necessity of recent factor. Besides, we also show the similarities among questions on three datasets in Figure 7. From figure 7, we can find there are many continue similarities among questions on the ASSISTments2009 and EdNet datasets. The proportion of continue similarities demonstrates the necessity of defining similarities as continuous values instead of binary values. Because many questions on Algebra2006 are related with one concepts and questions seldomly have continuous similarities.

## 4.2 Baselines

To validate the effectiveness of MF-DAKT, we compare our model against the state-of-the-art KT models.

*4.2.1 Deep Sequential Models*

- **DKT**. It's the first model which introduces deep learning methods into knowledge tracing model and uses a single layer LSTM model to predict students' performance.
- **DKVMN**. It uses a key-value memory network to extend DKT where the key matrix represents the relations between different skills and value matrix stores mastery of students.
- **DKT-Q**. It's an extension to the original DKT in which the questions are the input of model instead of skills.
- **DKVMN-Q**. It's also an extension to the original DKVMN and directly predicts students' response based on questions.
- **GKT**. It randomly builds a knowledge graph, and the built skill graph is used to predict. At the same time, the edge of graph can be optimized based on the data-driven method.
- **GIKT**. It uses a graph convolution network to capture the relations between questions and adopts a recap module to capture the long-term dependencies.
- **AKT**. It couples flexible attention-based neural network models with a series of novel and uses IRT model to avoid using an excessive number of parameters.

*4.2.2 Factor Analysis Models*

- **IRT**. It's a classical model in the knowledge tracing field and the simplest model for factor analysis. It only uses two parameters to respectively represent the ability of students and the difficulty of questions. Its function is as below:
$$y_{ij} = \theta_i - d_j \qquad (22)$$
- **MIRT**. It's the multidimensional version of the IRT model.
- **PFA**. It regards the positive and negative attempts as two situations to offer higher sensitivity to dynamic learning.

**Table 2: The Results over Three Datasets**

| Methods | ASSIST2009 | | Algebra06 | | EdNet | |
|---|---|---|---|---|---|---|
| | ACC | AUC | ACC | AUC | ACC | AUC |
| DKT | 0.721 | 0.744 | 0.838 | 0.781 | 0.656 | 0.692 |
| DKVMN | 0.728 | 0.751 | 0.842 | 0.786 | 0.651 | 0.687 |
| DKT-Q | 0.705 | 0.721 | 0.836 | 0.760 | 0.638 | 0.672 |
| DKVMN-Q | 0.731 | 0.753 | 0.844 | 0.791 | 0.662 | 0.708 |
| GKT | 0.692 | 0.726 | 0.859 | 0.743 | 0.657 | 0.689 |
| GIKT | 0.741 | 0.781 | 0.849 | 0.802 | 0.709 | 0.748 |
| AKT | 0.754 | 0.802 | 0.853 | 0.821 | 0.705 | 0.731 |
| IRT | 0.729 | 0.761 | 0.841 | 0.768 | 0.673 | 0.715 |
| MIRT | 0.731 | 0.760 | 0.842 | 0.770 | 0.675 | 0.721 |
| AFM | 0.671 | 0.624 | 0.836 | 0.699 | 0.634 | 0.641 |
| PFA | 0.708 | 0.705 | 0.840 | 0.741 | 0.638 | 0.655 |
| KTM | 0.751 | 0.792 | 0.845 | 0.794 | 0.685 | 0.730 |
| DAS3H | 0.743 | 0.785 | 0.842 | 0.793 | 0.683 | 0.731 |
| AKT-DAKT | 0.762 | 0.815 | 0.855 | 0.832 | 0.706 | 0.759 |
| PEBG-DAKT | 0.769 | 0.821 | 0.856 | 0.835 | 0.698 | 0.762 |
| **MF-DAKT** | **0.793** | **0.851** | **0.862** | **0.844** | **0.710** | **0.776** |

- **KTM**. It's the latest factor analysis model which uses the Factorization Machine (FM) [28] to model multiple factors. It covers many factor analysis models like IRT and AFM.
- **DAS3H**. It sets a time window to count students' attempts and uses KTM method to model students' learning progress.
- **AKT-DAKT**. We use IRT-based question representation method used in AKT. For fair comparison, we also firstly pre-train IRT-based question representations.
- **PEBG-DAKT**. We use PEBG method [17] to pre-train question representations and use DAKT to predict.

All methods are learned by optimizing cross-entropy loss using the Adagrad algorithm [8] except KTM and DAS3H which use MCMC [32]. For all baselines, we follow their work setting and source codes. The learning rate of our model is set to 0.001 and batch size for three datasets is set to 2048. We set embedding size $D$ as 128 which is common in KT researches [17], [40]. $\lambda_1$ and $\lambda_2$ in the pre-training process all are set to 0.5. We performed 5-fold cross validation. For each fold, 20% of students' records are held out as test set and remaining 80% are used as training set. The code is available at https://github.com/zmy-9/MF-DAKT.

## 4.3 Results and Discussions

In this section, we list the average test results of evaluation from 5 trials in Table 2 and the highest performance are denoted in bold. We can find the proposed model MF-DAKT achieves the highest prediction accuracy on all datasets. On the ASSSISTments 2009 dataset, MF-DAKT model achieves AUC of 85.1%. Among factor analysis models, KTM produces an average AUC of 79.2% which is the highest. As the state-of-the-art models, AKT and GIKT produces an AUC value of 80.2% and 78.1% respectively. Other models all perform worse. On Bridge Algebra2006 dataset, the AUC of MF-DAKT is 84.4%, which is better than 79.4% for KTM, 82.1% for AKT and 80.2% for GIKT. With regard to EdNet dataset, AUC of MF-DAKT is 77.6% and KTM gains the AUC of 73.1% and GIKT obtains an AUC of 74.8%. We also find PEBG-DAKT performed worse than MF-DAKT. We think the reason is that PEBG uses binary relations, and it's not reasonable to regard two concepts as related if they appear in one question.

**Table 3: Ablation Study of Key Components**

| Methods | ASSIST2009 | | Algebra06 | | EdNet | |
|---|---|---|---|---|---|---|
| | ACC | AUC | ACC | AUC | ACC | AUC |
| MF-DAKT | **0.793** | **0.851** | **0.862** | **0.844** | **0.710** | **0.776** |
| R-Recent | 0.789 | 0.841 | 0.854 | 0.823 | 0.708 | 0.769 |
| R-Pre | 0.785 | 0.838 | 0.852 | 0.816 | 0.694 | 0.747 |
| R-Reg | 0.786 | 0.841 | 0.856 | 0.832 | 0.698 | 0.763 |
| MF-KTM | 0.783 | 0.835 | 0.857 | 0.836 | 0.706 | 0.765 |

**Table 4: Ablation Study of Pre-training**

| Methods | ASSIST2009 | | Algebra06 | | EdNet | |
|---|---|---|---|---|---|---|
| | ACC | AUC | ACC | AUC | ACC | AUC |
| MF-DAKT | **0.793** | **0.851** | **0.862** | **0.844** | **0.710** | **0.776** |
| R-Relation | 0.787 | 0.836 | 0.858 | 0.836 | 0.703 | 0.769 |
| R-Diff | 0.784 | 0.835 | 0.853 | 0.822 | 0.702 | 0.766 |
| R-Binary | 0.773 | 0.824 | 0.857 | 0.838 | 0.690 | 0.753 |

**Table 5: Ablation Study of ACNN Component**

| Methods | ASSIST2009 | | Algebra06 | | EdNet | |
|---|---|---|---|---|---|---|
| | ACC | AUC | ACC | AUC | ACC | AUC |
| MF-DAKT | **0.793** | **0.851** | **0.862** | **0.844** | **0.710** | **0.776** |
| R-CNN | 0.785 | 0.839 | 0.859 | 0.840 | 0.707 | 0.770 |
| R-Attention | 0.784 | 0.841 | 0.858 | 0.839 | 0.706 | 0.768 |
| R-ACNN | 0.781 | 0.832 | 0.858 | 0.836 | 0.694 | 0.761 |

**Table 6: Ablation Study of DAKT Component**

| Methods | ASSIST2009 | | Algebra06 | | EdNet | |
|---|---|---|---|---|---|---|
| | ACC | AUC | ACC | AUC | ACC | AUC |
| MF-DAKT | **0.793** | **0.851** | **0.862** | **0.844** | **0.710** | **0.776** |
| R-Interaction | 0.761 | 0.811 | 0.857 | 0.834 | 0.701 | 0.764 |
| R-Feature | 0.786 | 0.837 | 0.855 | 0.825 | 0.692 | 0.763 |
| R-Dual | 0.776 | 0.825 | 0.858 | 0.837 | 0.701 | 0.765 |
| R-Attention | 0.786 | 0.832 | 0.858 | 0.838 | 0.704 | 0.767 |
| R-DNN | 0.789 | 0.838 | 0.856 | 0.840 | 0.702 | 0.758 |

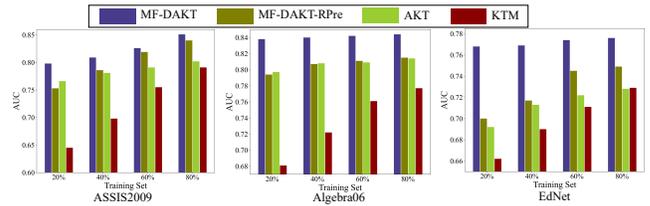

**Figure 8: Performance with different training set size**

From results listed in Table 2, we find DKT has the similar performance with DKVMM, but factor analysis models like IRT also achieve similar or even better performance than DKT which can show the efficacy of the traditional factor analysis methods. Moreover, we also find DAS3H can't perform better than KTM. We think the reason is that although DAS3H aims to model students' forget behavior, it introduces a lot of window-related parameters to be learned and we hardly selected a proper window length, leading to DAS3H performing worse than KTM.

**Training set size**. AKT points out the number of questions can cause overparameterization. Therefore, AKT uses parameters of the same scale as the number of concepts to represent questions based on IRT model [11]. To validate the robustness of MF-DAKT and dual-subspaces whether lead to overparameterization, we conduct experiments with different training set sizes: 20%, 40%, 60% and 80% of the total records. The results of MF-DAKT, AKT and KTM are shown in Figure 8. MF-DAKT-RPre means we remove the pre-training process. From Figure 8, we find that MF-DAKT can achieve excellent performance using only 20% of training data to train model. The performance of both MF-DAKT and AKT fluctuate within a small range which demonstrate both pre-training in MF-DAKT and IRT-based method in AKT can relieve overparameterization. We input question representations based on IRT method as AKT did into DAKT to predict students' response, i.e., AKT-DAKT which perform worse than MF-DAKT as Table 2 shows. We think reason is although IRT-based method can reduce parameters, it hardly captures the same information as each question owning an independent representation.

### 4.4 Ablation Study

In this section, in order to get deep insights on the contributions of various components in MF-DAKT, we also conduct some ablation prediction outcomes. We define some variations of MF-DAKT which denote missing one or more parts of the model. Specifically, R-Recent, R-Pre and R-Reg denotes we remove the recent factor, pre-training part (randomly initialization) and question difficulty regularization term, respectively. MF-KTM refers that DAKT is replaced with KTM model. Except for changes mentioned above, other components remain unchanged. Results are listed in Table 3.

As Table 3 shows, we find MF-DAKT performs the best on all datasets which demonstrate the effectiveness of each component of our proposed model. If we remove recent factor, R-Recent will perform worse especially on Algebra2006 dataset. The reason can be that the proportion of non-zero time interval on Algebra2006 is larger than other two datasets as Figure 5 and Table 1 show. Without pre-training process, we will use two randomly initialized representation matrixes and performance of R-Pre also degrades on three datasets. R-Pre's performances degrade the most on the Algebra2006 datasets which has 129, 225 questions and are larger than other datasets. The results validate the necessity of enriching question representations. Besides, R-Reg performs worse than MF-DAKT. We think the reason is that although pre-training can incorporate question difficulty level, a binary cross-entropy loss may make question representations lose difficulty information during the process of fine-tuning. At last, the result of MF-KTM validates the contribution of the dual-attentional mechanism.

What's more, in order to get more insights about our model, we separately conduct ablation studies on the pre-training, ACNN and DAKT components proposed in our model. In Table 4, R-

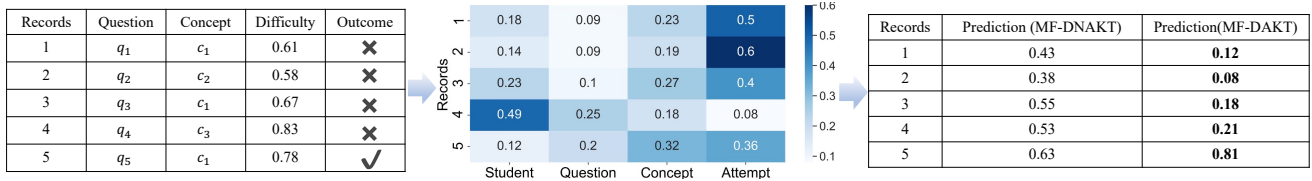

**Figure 9: Attention scores of factors in a student's practice records**

Relation and R-Diff separately mean we remove the relationship and difficulty level to pre-train. R-Binary denotes we set the similarities of questions as binary values. If we remove one of question information, model's performance will degrade. Besides, R-Binary also performs worse. Algebra2006 dataset has few questions involve multiple concepts, and R-Binary has the least performance degradation on Algebra06 dataset.

Table 5 and Table 6 separately shows the results of ablation studies of ACNN and DAKT components. R-ACNN in Table 5 means we concatenate all factors and directly input them into DAKT model. R-Interaction and R-Feature in Table 6 denotes we remove the factor interactions and factor subspaces, respectively. R-Dual in Table 6 means we project factors into a representation space to capture information. In Table 6, we can find if we only use one representation space, the performance will degrade which demonstrate the efficacy of utilizing two subspaces.

### 4.5 Visualization

In this section, we conduct some visualizations to get more deep insights into pre-training process and dual-attentional framework.

**Case Study.** We choose an example of a student's practices in EdNet and show corresponding attention weights outputted by our model in Figure 9, where attempt means the generated attempt vectors by ACNN. This student firstly answered question $q_1$ with difficulty level 0.61 which shows $q_1$ is not very difficult to solve by practicing it. Therefore, the attention score of attempt is the highest which means attempt has the most important contribution to prediction in the record 1. When we predict his record of 4, the target question $q_4$ has a high correctly answered rate. Therefore, our model predict record 4 mainly depend on students' ability. That' why the student factor has the highest attention score in prediction of this record. When student practiced question $q_5$, we find $q_5$ involves concept $c_1$ which he had ever practiced twice in record 1 and 3. Therefore, both the representations of concept factor and attempt vector get high attention scores during prediction process. At the same, we also list prediction outcomes on students' five records from MF-DAKT and MF-DNAKT in the most right part of Figure 9. MF-DNAKT means we remove attention mechanism. From Figure 9, we can find the predictions of MF-DAKT can be closer to true outcomes than predictions of MF-DNAKT by assigning factors unique attention scores.

What's more, we also use t-SNE [18] to visualize the question representations generated by pre-training and further optimized by DAKT in factor subspace on ASSISTments2009 dataset in Figure 10, where questions involving the same concepts are labeled in the same color. Figure 10. (a) shows the question representations from pre-training and we can find they are well-structured. After being further fine-tuned by DAKT, the question representations are projected into Figure 10. (b) which can show representations still retain the relationship and even are more well-structured than Figure 10. (a). We think the reason is that students' practices records can also help the pre-trained question representations to capture relations between questions during the fine-tune process. It's also the initial motivation for the design of GKT model [21].

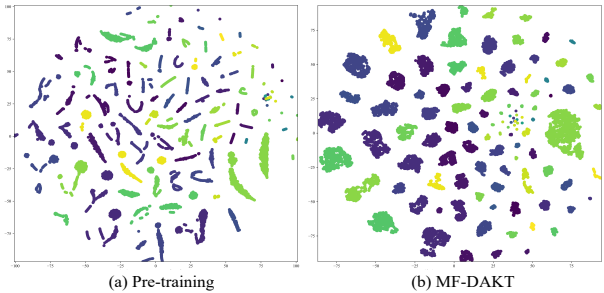

**Figure 10: Question Relation on ASSISTments2009 dataset**

## 5 CONCLUSION

In this work, we proposed a novel KT model to make up for the limitations of existing factor analysis methods. We propose a new factor to record the most recent relevant practices of students to highlight the impact of the most recent practices of students. We adopt a pre-training method to enrich question representations by rich question information including questions' relationship and difficulty level. Moreover, we apply a dual-attentional mechanism to project each factor into two representation subspaces so as to differentiate contributions of factors and factor interactions and fully capture information contained in data from two perspectives. At last, extensive experimentation results on three real-world datasets validate effectiveness of our model and each component.

In the future, we plan to collect text information of questions and combine with question graph proposed in this paper to further enrich question representations. Besides, we also try to introduce attempts on questions to model students' learning progress instead of only utilizing the attempts on concepts.

### ACKNOWLEDGMENTS
This work is supported by the Beijing Education Reform Project "Research and Practice of AI-Driven New Cultivation Mode for students majored in Information and Communication".